\documentclass[11pt]{article}

\usepackage{authblk}
\usepackage[T1]{fontenc}

\usepackage{acl2015}
\usepackage{times}
\usepackage{url}
\usepackage{latexsym}
\usepackage[acronym]{glossaries}

\usepackage{amsmath}
\usepackage{amssymb}
\usepackage{graphicx}
\usepackage[utf8]{inputenc}

\newacronym{GRASSHOPPER}{GRASSHOPPER}{Graph Random-walk with Absorbing StateS that HOPs among PEaks for Ranking}
\newacronym{LDA}{LDA}{Latent Dirichlet Allocation}
\newacronym{LSA}{LSA}{Latent Semantic Analysis}
\newacronym{POS}{POS}{Part-Of-Speech}

\title{Generation of Multimedia Artifacts: An Extractive Summarization-based Approach}

\author[1,2]{\rm Paulo Figueiredo}
\author[1,3]{\rm Marta Aparício}
\author[1,2]{\rm David Martins de Matos}
\author[1,3]{\rm Ricardo Ribeiro}
\affil[1]{L2F - INESC ID Lisboa, Portugal}
\affil[2]{Instituto Superior Técnico, Universidade de Lisboa, Lisboa, Portugal}
\affil[3]{Instituto Universitário de Lisboa (ISCTE-IUL), Lisboa, Portugal}
\affil[ ]{\textit {\{paulo.figueiredo,marta.aparicio,david.matos,ricardo.ribeiro\}@l2f.inesc-id.pt}}

\begin{document}

\maketitle
\begin{abstract}

We explore methods for content selection and address the issue of coherence in the context of the generation of multimedia artifacts. We use audio and video to present two case studies: generation of film tributes, and lecture-driven science talks. For content selection, we use centrality-based and diversity-based summarization, along with topic analysis. To establish coherence, we use the emotional content of music, for film tributes, and ensure topic similarity between lectures and documentaries, for science talks. Composition techniques for the production of multimedia artifacts are addressed as a means of organizing content, in order to improve coherence. We discuss our results considering the above aspects.

\end{abstract}

\section{Introduction}

We focus on the automatic generation of video artifacts, having as main concerns content selection and coherence aspects. Automatic video generation has been explored in a wide variety of areas, including generation of film trailers~\cite{Brachmann2007}, conference video proceedings~\cite{Amir:2004:AGC}, sports~\cite{Mendi:2013:SVS}, music~\cite{HUA:2004AMG}, and matter-of-opinion documentaries~\cite{Bocconi:2008:AGM}. 

In our work, specifically, as usual in text-to-text generation, the creation process is guided by an original document, the input. In that sense, the sequence of segments that composes a video artifact should correspond to parts that make-up a narrative following some intent~\cite{narrative:1992}. Content selection is driven by the text stream corresponding to the subtitles of the input document. Coherence is addressed by considering the audio and video streams.  To showcase our approach, we generate multimedia artifacts for two distinctly purposed case studies: film tributes; and lecture-driven science-talks.

For films, lectures, and documentaries, we use subtitles on account of their availability. We use timestamps to map text to the audio/video stream. All subtitles were segmented at the sentence level. Timestamps and punctuation inside sentences were removed. Specifically for films, we decided to use subtitles for the hearing-impaired due to their resemblance to scripts, which have been shown to produce better results in summarization tasks~\cite{MartaPaulo2015_1}.

Our first case study uses a film and song as input. The intended output artifact consists of a video containing important parts of the film along with the specified song.
Music is known to have a profound effect on humans' emotions~\cite{Picard:1997:AC}. For this reason, it is often used along with stories in order to emphasize their emotive content. In the same way, we address coherence by synchronizing the emotive content of the song with relevant, emotionally-related, parts of the film.  In order to maintain some consistency concerning the emotions portrayed by the content selected, centrality-based summarization approaches may be indicated. We use Support Sets~\cite{Ribeiro2011} as a ranking algorithm to determine the most central (important) sentences from the film, and let the emotions of the input song govern which content is presented.

Our second case study concerns video lectures of physics.
For this artifact, we use scientific documentaries to illustrate the main subjects addressed in the input lecture.
In contrast with our first case study, we produce a video from objective and informative sources. Lectures are talks structured around a syllabus that present information concerning a specific issue with a set of topics. Specifically, physics courses study real world phenomena that can be exemplified in various ways. 
We focus on obtaining a diverse representation of the lecture comprising different topics. For this reason, we summarize its subtitles using GRASSHOPPER~\cite{Zhu2007}, which maximizes diversity while penalizing redundant content. We obtain a thematically-coherent artifact by selecting topic-related content from a collection of documentaries.

This article is organized as follows:   Section~\ref{sec:related_work} presents related work for video generation; Section~\ref{sec:FilmTribute} presents our first case study: film tributes. Section~\ref{sec:sciencetalks} presents our second case study: lecture-driven science-talks.   Section~\ref{sec:discussion} presents a discussion of our early results; Section~\ref{sec:conclusions} presents the conclusions and directions for future research.

\section{Related Work}\label{sec:related_work}

In the following sections, we present work concerning content selection and coherence aspects for automatic generation of multimedia artifacts.

\subsection{Content Selection}

It is essential to extract important content in order to generate a video that raises the viewer's interest. For example,~\newcite{Ma:2002:UAM} present a method that models the user's attention in order to create video summaries. However, despite the numerous techniques to produce video skims~\cite{Truong:2007:VAS}, results are still far from human expectations. Many approaches neglect the audio, due to the difficulty of integrating its features in video~\cite{Yingbo:2014:MMMRMV}. Other approaches, propose the fusion of text, audio, and visual features, with relation to particular topics, for multimedia summarization~\cite{Ding:2012:BAV}.
Given the availability of subtitles for films, lectures, and documentaries, text can be exploited to determine content.~\newcite{sumFilm} summarize films using its subtitles (text), along with information from the audio and the visual streams, integrating cues from these sources in a multimodal saliency curve. Auditory saliency is determined by cues that compute multifrequency waveform modulations, visual saliency is calculated using intensity, color, and orientation values, and textual saliency is obtained through \gls{POS} tagging. 
Several generic summarization algorithms have been developed to determine relevant content, for instance methods based on: centrality~\cite{Erkan2004,Ribeiro2011}; diversity~\cite{Carbonell1998,Zhu2007}; or uncovering latent structure~\cite{Gong2001}.

\subsection{Coherence}

Extractive summaries are known to lack coherence~\cite{Paice:1990}.~\newcite{Foltz:LSAC} proposed an automatic method that addresses coherence as semantic relations between adjacent sentences. \gls{LSA}~\cite{landauer1998introduction} is used to uncover latent structure of the input, then sentences are represented as vectors, composed by the means of the words they contain. Computing the similarity between sentences determines aspects of local coherence.
Specifically, for video generation, coherence can be established by means of composition techniques for video production, based on temporal constraints, along with thematic and structural continuity~\cite{AhangerL:98}. Additionally, music data can be explored as a mechanism to provide coherence.
\newcite{DBLP:conf/mm/WuXQT12} produced a music video composed by web images and a song. The song lyrics were used to search for images in the web, which compose the video, based on estimated semantic scores between an image and a music segment.
\newcite{trailer} generate a film trailer that selects the most emotional film segments, and re-order the resulting set of shots so as to optimize the shot sequence, based on each shot emotional impact.

\section{Case Study 1: Film Tribute}\label{sec:FilmTribute}

\begin{figure}
	\begin{center}
		\includegraphics[keepaspectratio=true,width=\columnwidth]{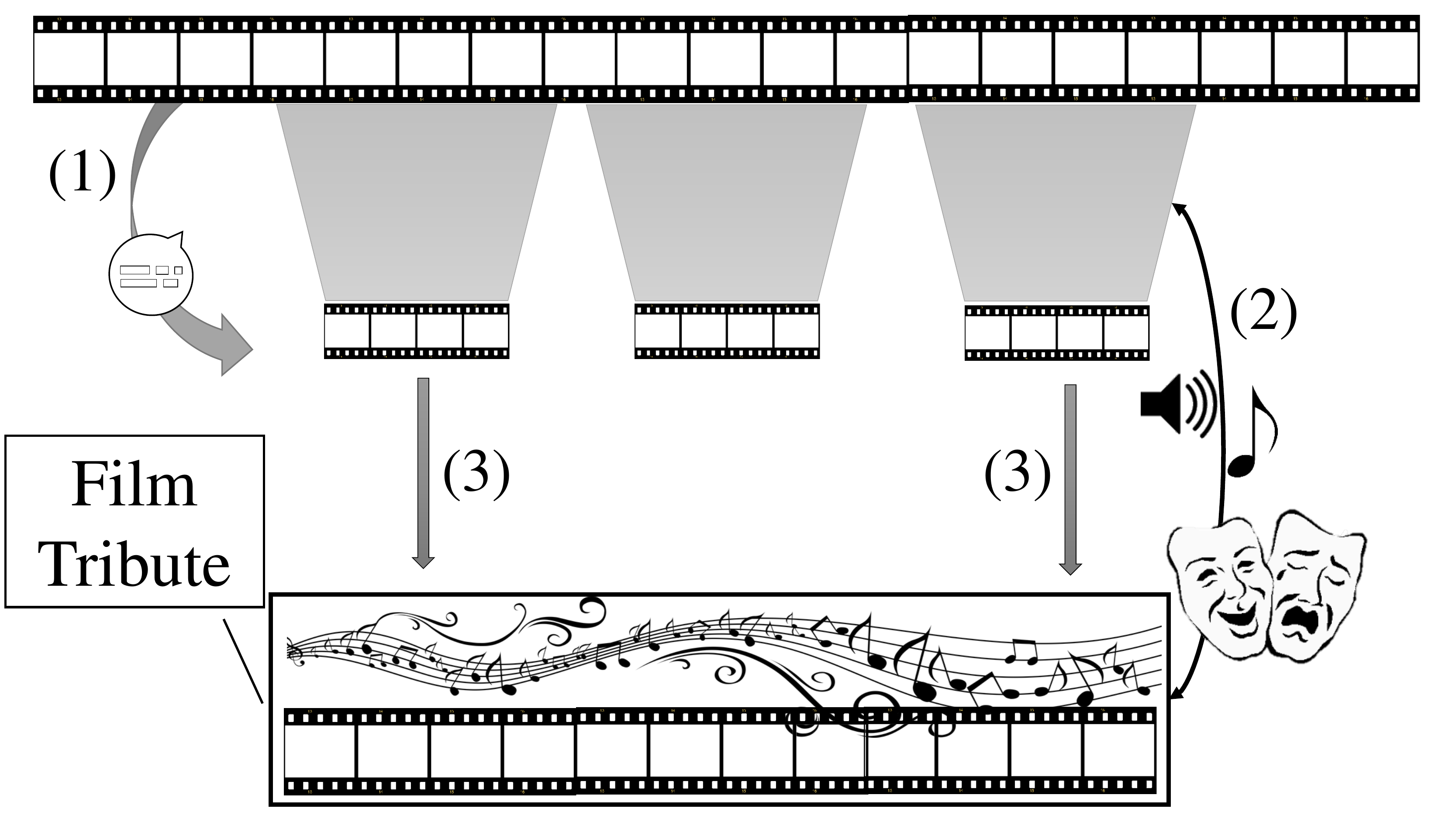}
	\end{center}
	\caption{Film tribute generation. (1) content selection. (2) emotion synchronization. (3) video composition.}
	\label{fig:goal}
\end{figure}

Figure~\ref{fig:goal} shows the processes involved in the production of a film tribute, given a film and a song as input. The length of the song imposes the duration of the final artifact, which is populated by content selected from the film. In order to maintain some emotional consistency concerning the  selected content, we use a centrality-based approach. For this reason, we obtain relevant sentences from the film, summarizing its subtitles, using Support Sets. This algorithm uncovers groups of semantically-related passages. A support set is created for each passage of the input, determined by comparing each passage with all remaining ones from the source. A summary is composed by the most relevant passages, which are the ones present in the highest number of support sets.
Furthermore, we use the subtitles timestamps to obtain the matching video clips. For each one, we detect the corresponding scene. Then, we extract emotion-related audio features from the music and the scenes of the video clips, and compare them to obtain the ones that are more emotionally similar to the music. The scenes provide more auditory information concerning the events involving the extracted video clip. 
The final video is composed by joining the resulting video clips with the specified music.

In order to obtain the film's scenes, we segment it using  Lav2yuv, a program distributed with the MJPEG tools~\cite{conf/aina/ChenSL12}, with the scene detection threshold fixed to 40. 
Then, we extract emotion-related audio features from the video scenes and  music, which include 384 features as statistical functionals applied to low-level descriptor contours (INTERSPEECH 2009 Emotion Challenge feature set from openSMILE~\cite{Eyben:2010:OMV}). The 16 low-level descriptors are the following:  

\begin{itemize}
	
	\item Root-mean-square signal frame energy;
	\item Mel-frequency cepstral coefficients 1-12;
	\item Zero-crossing rate of time signal (frame-based);
	\item The voicing probability computed from the ACF;
	\item The fundamental frequency computed from the Cepstrum.
\end{itemize}

The resulting vector for each clip is then compared with the music vector using the cosine distance. If the similarity between them is greater than 0.7 (empirically-determined value), we consider that the video clip has the same emotion of the music. 
The length of the audio clip containing the music is filled with the resulting clips. 
The final video is composed by joining the resulting video clips with the specified music, following the chronological order of the input film.

\section{Case Study 2: Lecture-Driven Science-Talks}\label{sec:sciencetalks}

\begin{figure}
	\begin{center}
		\includegraphics[keepaspectratio=true,width=\columnwidth]{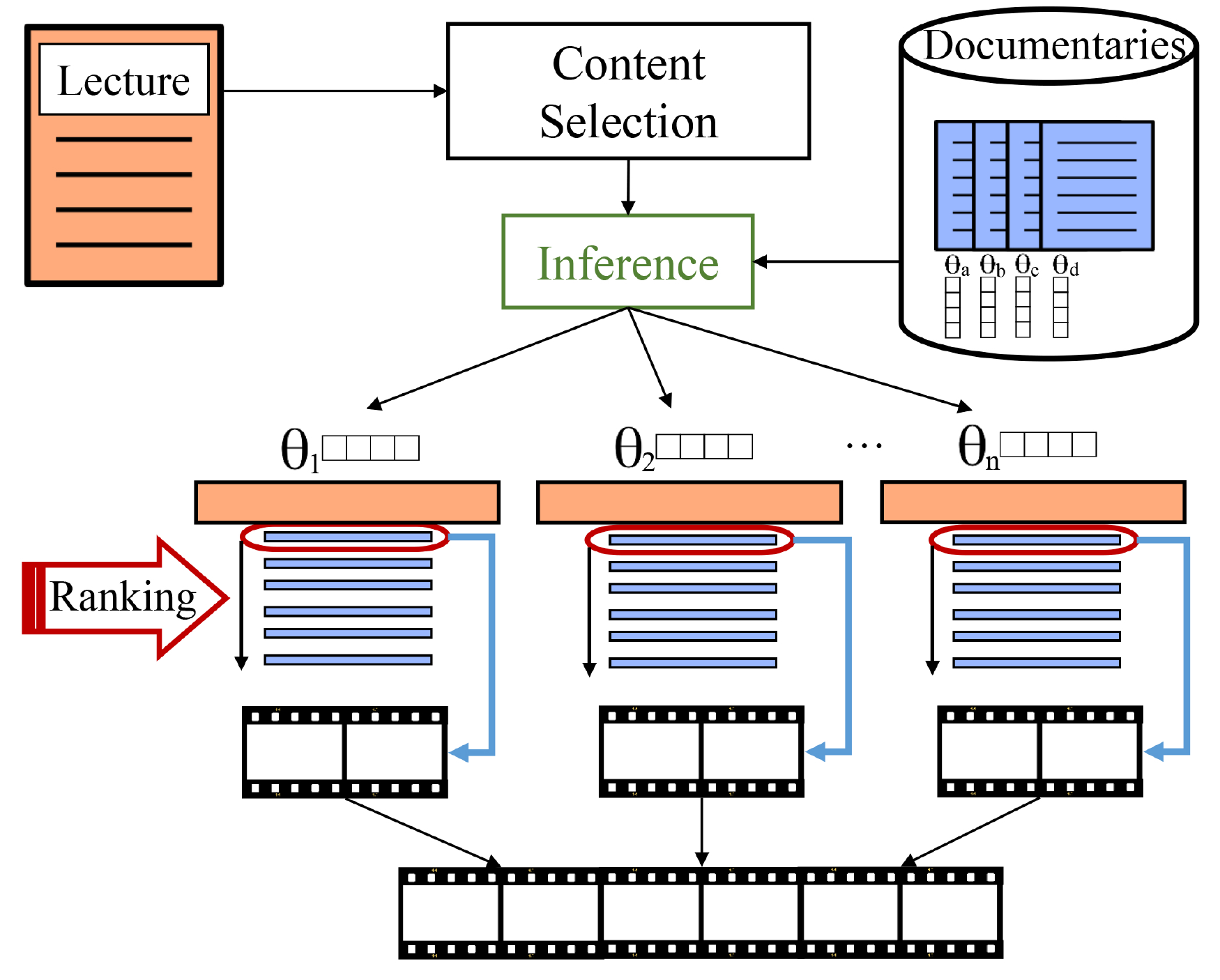}
	\end{center}
\caption{Lecture-driven science-talks video generation architecture: $\theta_i$ represent topic mixtures.}
	\label{fig:videogeneration}
\end{figure}

Figure~\ref{fig:videogeneration} depicts the architecture of the computational method developed for the generation of lecture-driven science-talks.  Our method receives as input the lecture's subtitles. First, we use GRASSHOPPER, a diversity-based approach, to obtain the most important, yet diverse, set of sentences from the lecture. GRASSHOPPER is a graph-based ranking algorithm based on random-walks in an absorbing Markov chain, which focuses on maximizing diversity while minimizing redundancy. The list of items returned by the algorithm is computed based on their representation of particular groups (centrality), the difference between them (diversity), and prior ranking.
Then, we determine, for each sentence, topic-related content from a collection of documentaries, as a means of obtaining thematically-coherent content. In the following sections, we detail two major approaches carried out using \gls{LDA}: (i) model trained at sentence-level; (ii) inferring the lecture in a model trained at document-level, in order to obtain a subset of topic-related documentaries, then training an additional model for this subset at sentence-level. Finally, we present the steps involved in choosing the best candidates to compose the final artifact.

\subsection{\gls{LDA}-Based Content Selection}\label{subsubsec:lda}

\gls{LDA}~\cite{Blei:2003:LDA} is a generative probabilistic model that explains document collections based on mixtures of topic distribution. In this sense,~\gls{LDA} follows the intuitive notion that documents exhibit multiple topics. Each topic is characterized as a distribution over words in a fixed vocabulary, and each document is a random mixture of topics. Given the trained model, is it possible to find related documents by assessing the similarity of their topic mixtures.

\subsubsection{Sentence-Level Model}\label{paragraph:sentence-level_model}

We train an~\gls{LDA} model for the collection of documentaries using 100 topics at sentence-level, uncovering the hidden thematic structure of the collection. Then, we situate new data, namely the summarized lecture, into the estimated model. Therefore, we fit the summarized lecture into the topic structure learned using variational inference. Now, the sentences of the summarized lecture and the collection of documentaries can be compared by looking at how similar their topic mixtures are.
Given that GRASSHOPPER aimed at maximizing the diversity of the condensed lecture, we posit that each sentence of the summary regards different topics. As a result, we compare, based on topic mixtures, each sentence of the lecture with sentences from documentaries, using the cosine distance, and obtain a top-$k$ containing topic-related candidate documentary sentences for each lecture's sentence. As a first approach, all experiments were made with $k$ fixed at 10.

\subsubsection{2-Stage Model}\label{paragraph:2-stage_model}

In this approach, we first train an~\gls{LDA} model for the collection of documentaries using 100 topics at document-level. Then, by fitting the lecture-document in the estimated model, we use its mixture topics  to obtain the most relevant documentary topic-related documents. This first step allows a focused determination of the candidate segments. After that, the procedure presented above is applied. The model is now trained at sentence-level, for the subset of documentaries using a smaller number of topics, for instance, 10. As before, the lecture is situated into this new model and the cosine distance determines the top-$k$ with topic-related candidate documentary sentences.

\subsection{Ranking}\label{subsubsec:ranking}

Given the top-$k$ candidates, we want to choose the sentence that best represents the group. For this reason, we use the Support Sets algorithm, in order to let sets of candidate sentences compete among them and, thus, obtaining a sentence ranking.
In our current work, the final video is composed by the best ranked candidate sentences of documentaries. However, since the same sentence can be present in more than one lecture segment, which would result in redundant content in the final artifact, only unused candidates are chosen (if the best ranked sentence was already selected, the next in rank is used).

\section{Results and Discussion}\label{sec:discussion}


We conducted preliminary experiments in order to incorporate the viewer's feedback into the generation process. Several issues were identified: (i) the pace at which the video progresses, the result of concatenating video clips that correspond to sentence-level segments; (ii) the text stream is mapped to the video using subtitles, occasionally, causing the time interval corresponding to the sentences of the subtitles not to encompass the speech that it is portraying; (iii) overlapping music with video segments can clutter the audio stream by making the video's speech unclear; and, (iv) when joining audio/video segments from different sources, the lack of loudness consistency in the final audio stream affects the overall user experience. In the following sections, we discuss our approaches for content selection and address the aspects of coherence in light of our early results.

\subsection{Content Selection}

In our work, content selection is driven by a text stream that corresponds to transcripts of speech monologues and dialogs, presented in the input document's subtitles. In that sense, we do not detect important content based on visual or audio cues, except those corresponding to speech (via subtitles). Hereof, other approaches can be used~\cite{Coldefy:TVAAVC,Coldefy:2004:USV}. For text-based selection, different approaches are available depending on the aspects of interest (such as diversity).

For film tributes, which target the viewer's emotions, algorithms that focus on the most central (important) content may be indicated. Apart from Support Sets, other algorithms can be used, for instance, LexRank~\cite{Erkan2004}, which is also a centrality-based algorithm. In contrast with film tributes, lecture-driven science-talks are instructive. Thus, in order to capture sentences related to the various topics in the lecture, other algorithms based on diversity can be used, for instance, MMR~\cite{Carbonell1998}, that provides a model that linearly combines relevance and novelty.

Our choices were based on previous work (omitted for blind review) that shows that Support Sets and GRASSHOPPER provide better summaries for the content selection phase.

\subsection{Coherence}

Regarding film tributes, our experiments show that viewers are able to establish correspondence between the emotions of the film and the accompanying music. However, to improve results, thematic coherence can be considered: if the song has lyrics, they can also be taken into account to relate its topics to the film.

Regarding lecture-driven science-talks, one of the drawbacks of sentence-to-sentence substitution is the lack of continuity and the perceived fragmentation of the final artifacts. Although topic analysis was used to address thematic coherence, our preliminary experiments show that it is not enough for a viewer to regard the final video as fluid and well-composed. Additionally, considering our two model approaches, the 2-stage model seems to be better than the sentence-level model: most of the content is provided by the subset of documentaries, the final video contains segments focused on less documentary episodes. However, this problem merits further study.

Overall, both methods use subtitles segmented at sentence-level. For film tributes, the song is used to obtain an emotionally-coherent multimedia artifact. Results show that segmentation at sentence-level does not affect, significantly, its overall coherence. In contrast, for informative videos, sentence organization in the final artifact is of critical importance: although thematic coherence can be easily identified, the displayed content progresses without causality through the video segments that compose the artifact, adversely affecting the viewer's overall experience. Furthermore, video segments obtained from the input subtitles usually have a specified duration longer than the corresponding speech audio segment. As a result, the video segments have abrupt transitions, sometimes with considerable color variations. This last effect was clearly pointed out as a limitation by viewers. We plan to address these issues in the near future.

\section{Conclusions and Future Work}\label{sec:conclusions}

We presented methods for generation of multimedia artifacts, focusing on content selection and coherence aspects. We produced two types of video: film tributes, and lecture-driven science-talks. Each case study considers a different creative intent. 
While film tributes intend to appeal to the emotive-side of the viewer, lecture-driven documentaries are expected to provide a more instructive experience. 
Although our preliminary experiments had a good audience feedback, further improvements regarding aspects of coherence still need to be made. 

Regarding future work and concerning the identified issues, in the case of film tributes, if the music has vocals, its lyrics can be taken into account to relate their topics to the film. A possible solution is to receive only the film as input, then, choose a topic-related song from an existing dataset, for instance, The Million Song Dataset~\cite{Bertin-Mahieux2011}. Furthermore, considering music with vocals, audio adjustments can be made, specifically, in moments where the music's energy is high enough to interfere with speech from the video.  Also, the final video's volume can be adjusted when the music's energy is higher than some threshold. Furthermore, to improve the final artifact's structural coherence, we can take into account the music's structure and align it to the video stream~\cite{OriolNietoJuanPabloBello}.

Still regarding coherence, we intend to identify locally-coherent sentences for each method's input process: the summarized film, and lecture. \gls{LSA} can be used as a technique for measuring coherence, by comparing vectors of adjacent sentences in the generated semantic space.  Thus, they can be considered as  groups of locally-coherent sentences. For films, these groups can be directly used. For lectures, we can identify the topic mixtures that best represent each group and replace them by choosing other groups of locally-coherent documentary sentences.

Video composition techniques for video production can be used as a means to build a narrative, which can be seen as a series of events in a chain~\cite{narrative:1992}.  To resolve the identified abrupt shot transitions, we can use the underlying audio stream to provide continuity cues. For that, a data-drive voice-activity detector based on Long Short-Term Memory Recurrent Neural Networks (LSTM-RNN) can be used~\cite{conf/icassp/EybenenWSS13}.
Furthermore, content progression in the final multimedia artifact can be established by comparing adjacent video segments and ensuring that they are not too similar or too different~\cite{AhangerL:98}.

\bibliographystyle{acl}
\bibliography{acl2015}

\end{document}